\documentclass[10pt,twocolumn,letterpaper]{article}
\usepackage[pagenumbers]{cvpr} 
\usepackage{derivative}
\usepackage{amsmath,amssymb}
\usepackage{comment}
\definecolor{cvprblue}{rgb}{0.21,0.49,0.74}
\usepackage[pagebackref,breaklinks,colorlinks,allcolors=cvprblue]{hyperref}

\newcommand{\ournet}{Y-MAP-Net} 

\title{Y-MAP-Net: Real-time depth, normals, segmentation, multi-label captioning and 2D human pose in RGB images}

\author{Ammar Qammaz\\
Institute of Computer Science, FORTH and\\ 
Computer Science Department, University of Crete\\
{\tt\small ammarkov@ics.forth.gr}
\and
Nikolaos Vasilikopoulos\\
Computer Science Department, University of Crete, and\\ 
Institute of Computer Science, FORTH\\
{\tt\small nvasilik@ics.forth.gr}
\and
Iason Oikonomidis\\
Institute of Computer Science, FORTH,\\
N. Plastira 100, Vassilika Vouton,\\
GR70013, Heraklion, Crete, Greece\\
{\tt\small oikonom@ics.forth.gr}
\and
Antonis A. Argyros\\
Institute of Computer Science, FORTH and\\ 
Computer Science Department, University of Crete\\
{\tt\small argyros@ics.forth.gr}
}

\begin{document}
\maketitle
\begin{abstract}
We present \ournet, a Y-shaped neural network architecture designed for real-time multi-task learning on RGB images. \ournet\, simultaneously predicts depth, surface normals, human pose, semantic segmentation, and generates multi-label captions—all from a single network evaluation. To achieve this, we adopt a multi-teacher, single-student training paradigm, where task-specific foundation models supervise the network’s learning, enabling it to distill their capabilities into a lightweight architecture suitable for real-time applications. \ournet\, exhibits strong generalization, simplicity, and computational efficiency, making it ideal for robotics and other practical scenarios. To support future research, we will release our code publicly.
\end{abstract}    
\section{Introduction}
\label{sec:intro}
Decades of research have yielded powerful methods that robustly tackle long standing problems for computer vision and pattern recognition. These so-called foundation models stand out for their exceptional generalization ability, achieved through immense scale and complexity. These models, with billions of parameters, are trained on vast datasets, enabling zero-shot problem-solving across diverse tasks and robust performance on in-the-wild data. However this sheer size comes at a cost: they require PetaFLOP-scale computational resources and data centers equipped with thousands of GPGPU accelerators for a single training session, hindering reproducibility. 
Finally, they are expensive to deploy and impractical for real-time applications.

Meanwhile, we observe a class of much sparser convolutional models, such as the successful family of You Only Look Once (YOLO)~\cite{redmon2016you} models, which are profoundly influential in many different real-world applications. This motivates us to develop a framework that bridges the gap between real-time, multifaceted scene understanding and high-fidelity dense output. Once provided, it can be used as a building block to solve other general problems, in vision, robotics, human-computer interaction, surveillance, etc, by providing a unified, streamlined source of multi-modal data.

\begin{figure}[t]
  \centering
   \includegraphics[width=0.99\linewidth]{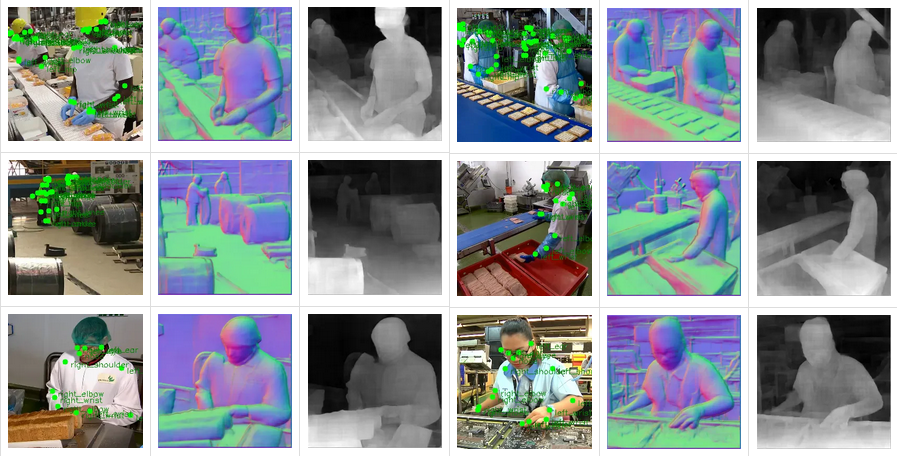}
   \caption{Given an RGB frame, \ournet~estimates human pose, depth, surface normals, segmentation and image captioning in real-time. The figure shows pose keypoints, depth and normal estimations on publicly available images from factory floors.}
   \label{fig:onecol}
\end{figure}

Given this motivation, in this work, we propose \ournet, a {\bf Y}-shaped {\bf M}ulti {\bf A}ttribute {\bf P}rediction neural network (NN) model that can be used as a closed-loop processor for RGB streams supplying human pose, depth, surface normals, image segmentation and a textual multiclass caption for each incoming frame (Fig.~\ref{fig:qualitative}).
To achieve this, we adopt a multi-teacher, single-student training paradigm~\cite{you2017learning,bruce2021multimodal}, 
where \ournet \, learns from task-specific foundation models supervising its predictions.

This approach distills large models into a compact architecture, enabling efficient multi-task learning without sacrificing generalization. In particular, \ournet \, accepts monocular RGB input and delivers 44 heatmap / image outputs and 8 caption tokens.\\
In summary, our main contributions are the following:
\begin{itemize}
  \item \ournet \, is the first convolutional NN method to achieve simultaneous depth, normal and human pose estimation while also providing scene segmentation and captioning from monocular RGB in a monolithic network.
  \item Due to its efficiency, the method we present performs in real-time and thus opens the way to many practical applications, especially in the field of robotics, complementing existing models for the various tasks we undertake.
  \item The proposed \ournet~ topology offers a novel, simple and pragmatic framework for multi-task learning.
\end{itemize}
\section{Related Work}
\label{sec:related}
The proposed method attempts to simultaneously tackle multiple mature sub-fields of the computer vision research landscape.  To achieve such a generalist technique we studied the various domains, identified their commonalities, best practices, useful data primitives and then gradually integrated them under a common unified formulation.

\noindent{\bf Image Classification and Captioning:}
AlexNET~\cite{krizhevsky2012imagenet} was the first method to successfully deploy a large scale convolutional neural network, train it  with a large dataset and perform 1000-way softmax classification. Although more than a decade has passed since this seminal work~\cite{alom2018history}, GPGPU acceleration and techniques like data augmentation, dropout, and ReLU activations have become mainstream and are used with small changes in our work.
The VGG network family~\cite{simonyan2014very} demonstrated the importance of increasing depth, with depth ranging from 11 to 19 layers, also leveraging $3 \times 3$ convolutions.
GoogLeNet~\cite{szegedy2015going} introduced an even deeper architecture featuring multiple data paths with different kernel sizes between layers, pioneering $1 \times 1$ convolution kernels. Finally, ResNETs~\cite{he2016deep} introduced residual connections that improved gradient flows throughout the network, allowing NNs to grow up to 1202 layers deep, dealing with the same classification task.
Meanwhile, works like DenseCap~\cite{Johnson_2016_CVPR} moved away from labels to dense text captioning while in the Natural Language Processing (NLP) field Word2Vec~\cite{mikolov2013efficient,mikolov2013distributed},  GloVe~\cite{pennington2014glove} and Fasttext~\cite{joulin2016bag} embeddings encoded text in a multi-dimensional representation with conceptual relations between vectors. The advent of transformers~\cite{vaswani2017attention}, a technique built from the ground up to capture textual relations, replaced other LSTM~\cite{6795963} or GRU~\cite{cho2014learning} based attempts. ConvCAP~\cite{Aneja_2018_CVPR} was the last major purely convolutional dense captioning work. Vision transformers (ViTs)~\cite{dosovitskiy2020image} inspired development of techniques like CLIP~\cite{radford2021learning} and DINO~\cite{caron2021emerging} that directly associated image and text embeddings, paving the way for state of the art VLLM foundation models like GPT4~\cite{openai2024gpt4technicalreport} and LLAMA 3.2~\cite{dubey2024llama3herdmodels} that successfully tackle the problem at its core.

\noindent{\bf Scene Segmentation:}
An alternative approach to scene captioning is annotating visible parts of the scene. YOLO~\cite{redmon2016you} served as a seminal work for image region classification. Building on this, Mask R-CNN~\cite{he2017mask} and Faster R-CNN~\cite{ren2016faster} advanced dense segmentation and per-pixel label classification, inspiring subsequent works like Mask YOLO~\cite{maskyolo2018}—fusing earlier approaches~\cite{he2017mask, redmon2016you}—as well as Detectron 2~\cite{wu2019detectron2}, DPText~\cite{ye2022dptext}, and Segment Anything~\cite{kirillov2023segment}.

\noindent{\bf Pose Estimation:} AlexNET~\cite{krizhevsky2012imagenet} directly inspired Human Pose Estimation (HPE) methods~\cite{toshev2014deeppose,tompson2014joint} that adapted it for joint heatmap regression. VGGs~\cite{yang2016end} and ResNETs~\cite{sun2017revisiting,luvizon2017learning}, followed by stacked hourglass networks~\cite{newell2016stackedhourglassnetworkshuman} were among the first new NN topologies for the HPE task. OpenPose was a landmark work that introduced part affinity fields (PAFs)~\cite{cao2017realtime}. During the same year, LCR-Net~\cite{rogez2017lcr} and Coarse-to-fine 3D human pose~\cite{Pavlakos_2017_CVPR} also made significant contributions in terms of 3D accuracy. DensePose~\cite{guler2018densepose} explored an alternative formulation of the problem by regressing UV-mapped coordinates to a 3D model, thus providing a much denser result than 3D joint positions. HRNET~\cite{sun2019deep} modified U-NETs~\cite{ronneberger2015u} by featuring a pyramid of different input resolutions and achieved new levels of accuracy. This inspired a host of more approaches~\cite{wang2020deep,cheng2020higherhrnet,yu2021lite,xu2022zoomnas}. The advent of ViTs~\cite{dosovitskiy2020image} shifted focus to transformer-based architectures~\cite{ci2023unihcp}, with the Sapiens~\cite{khirodkar2024sapiens} foundation model being the first to achieve 2D joint, depth and normal estimation, as well as part segmentation but focusing only on humans.

\noindent{\bf Depth and Normals Estimation:} MiDaS~\cite{ranftl2020towards} was a major push towards monocular depth estimation showcasing the power of relative depth. ZoeDepth~\cite{bhat2023zoedepth}, Metric3D~\cite{yin2023metric3d} and Marigold~\cite{ke2023repurposing} further refined accuracy, while Dust3R~\cite{wang2024dust3r}, UniDepth~\cite{piccinelli2024unidepth} and DepthAnything (DA)~\cite{yang2024depth} scaled in both model size and training data leading to DAv2~\cite{depth_anything_v2} a mature framework we use as the teacher for our depth training. 

\noindent{\bf Training Datasets:} A wide variety of data sources like LAION-5B~\cite{schuhmann2022laion}, Instagram-3.5B~\cite{mahajan2018exploring}, JFT-300M~\cite{sun2017revisiting}, and V.Genome~\cite{krishna2017visual}, take advantage of big-data effectiveness in deep learning, providing enormous training corpora, especially fitting for image captioning. Humans-300M~\cite{khirodkar2025sapiens}, Panoptic~\cite{joo2015panoptic}, AMASS~\cite{mahmood2019amass} and Motion-X~\cite{lin2024motion} are the largest HPE datasets.
Common Objects in Context (COCO)~\cite{lin2014microsoft} remains a balanced middle-ground for captioning, scene segmentation and HPE. It has sufficient size and scope to facilitate reproducible research without being overly encumbering in terms of required training resources.
Our approach stands on the ``shoulders of giants'' of prior research, drawing from a substantial body of computer vision work. At its core, our NN topology has similarities to U-NETs~\cite{ronneberger2015u}, however being pushed to new limits in-terms of multiple concurrent tasks.
Although Y-shaped architectures featuring two output branches have been proposed to tackle specialized medical applications~\cite{mehta2018net,wang2019net}, our \ournet~ formulation is novel by providing captions output using GloVe~\cite{pennington2014glove} embeddings.
In contrast to other caption works like ConvCAP~\cite{Aneja_2018_CVPR}, it does not iteratively query the network for each captioning token. Additionally, our model simultaneously regresses 2D joint heatmaps and PAFs, inspired by OpenPose~\cite{cao2017realtime}, and predicts depth and surface normals supervised by DepthAnything~\cite{yang2024depth}, with conceptual parallels to DensePose~\cite{guler2018densepose}. Moreover, supervised by Detectron 2~\cite{wu2019detectron2} and DP-Text~\cite{ye2022dptext} our work is equipped with per pixel object and scene segmentation capabilities.

A method with multi-modal output and conceptual similarities to 
\ournet, though not directly comparable, is the Sapiens~\cite{khirodkar2025sapiens} foundation model. Unlike our method, Sapiens focuses solely on the human body, not producing results for the entire input image. While Sapiens uses a 500M training corpus, our approach relies on just 0.1M samples. Architecturally, Sapiens is transformer-based, whereas we adopt a convolutional approach.
Instead of Sapiens' separate encoder models for each modality~\cite{sapiens_models}—ranging from 0.3B to 2.0B parameters—our monolithic model is $4\times$ smaller than the aggregate size of the minimum Sapiens setup to tackle all tasks. In our monolithic model, weights are shared between tasks, in contrast to the sequential approach adopted in Sapiens. Finally, we include captioning on the same model, in contrast to Sapiens~\cite{khirodkar2025sapiens} which lacks this functionality. All these differences render our approach not directly comparable to Sapiens.
\begin{figure}[t]
  \centering
   \includegraphics[width=1.0\linewidth]{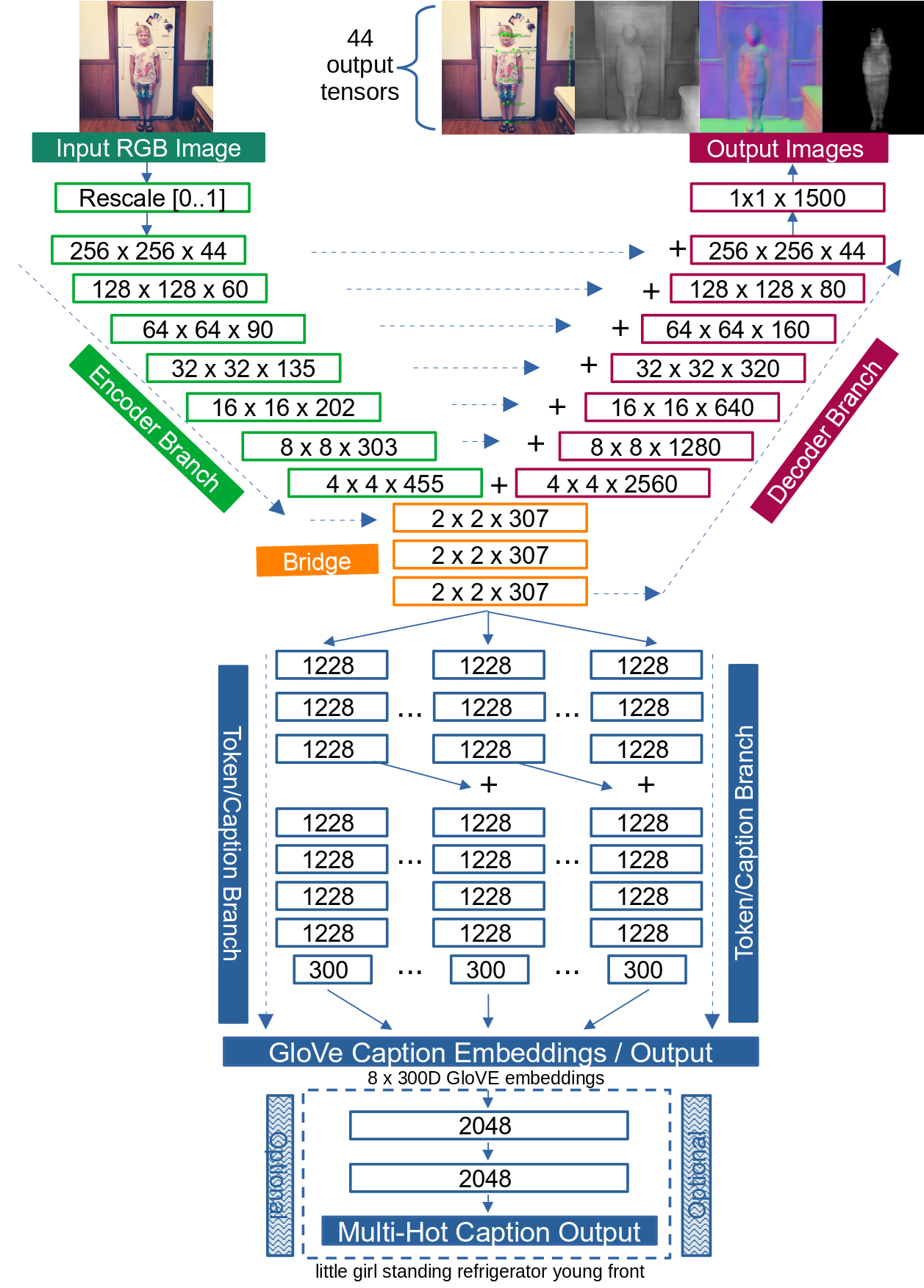} 
   \caption{ Architecture of the proposed \ournet. The flow of data from input to outputs is highlighted with cyan arrows. Residual connections are indicated with + signs, while rectangles give a dimensionality overview for each layer. Green layers (top left) signify encoder blocks originating from 1 RGB image. The network bridge layers appear in orange color (middle). Blue layers (bottom) depict 8 captioning output token GloVe~\cite{pennington2014glove} vectors. These can be optionally converted to multi-hot labels with the addition of 2 dense layers for some applications. Magenta (top right) signifies decoder blocks that lead to 44 multi-modal outputs. Detailed encoder and decoder layer architecture is provided in Figure~\ref{fig:encdec}.
   }
   \label{fig:model}
\end{figure}

\section{\ournet~ Model Architecture}
\ournet~ has a single {\bf input} or {\bf encoder} branch, marked with green in Figure~\ref{fig:model}. Its outputs diverge in two dedicated paths.
The first, marked with magenta, is related to {\bf pictorial} or {\bf decoder output}: depth, normals, heatmaps, PAFs, and segmentation masks.
The second, marked with blue, is related to {\bf token/caption output}.

These two paths are separated at the {\bf bridge} section (orange color in Fig.~\ref{fig:model}).
This structure effectively combines a U-NET-like~\cite{ronneberger2015u} architecture for pixel-level related tasks with a densely connected token encoder ensemble for caption generation.

\begin{figure}[t]
  \centering
   \includegraphics[width=1.0\linewidth]{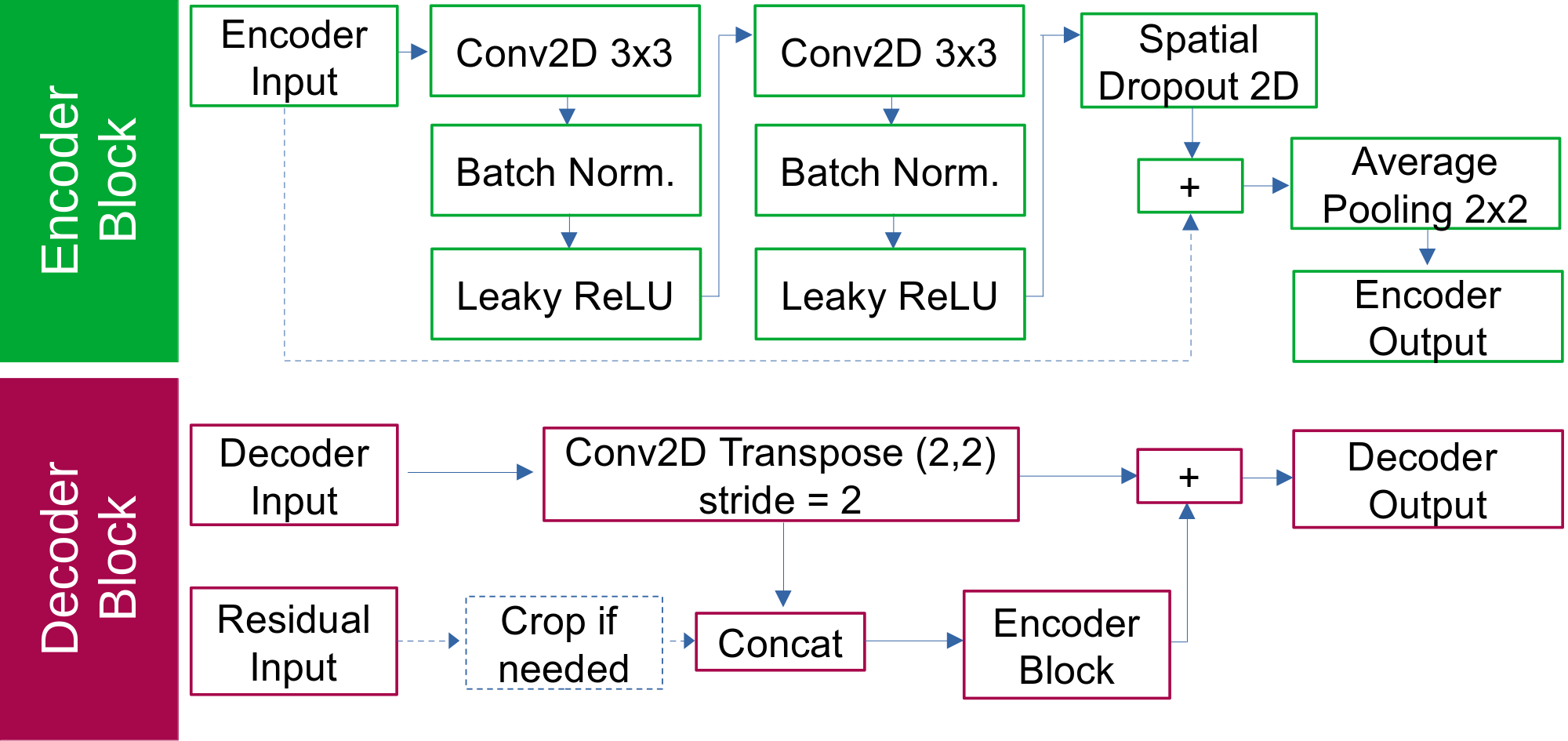} 
   \caption{ Encoder (green) and decoder (magenta) blocks of \ournet~ (Fig.~\ref{fig:model}) are tasked with down-scaling input images to the bridge representation and then up-scaling to pictorial outputs. They consist of layers and skip connections shown in this figure. }
   \label{fig:encdec}
\end{figure}

\noindent{\bf Encoder branch:} This branch  gradually distills lower dimensional feature embeddings until arriving at the bridge section. This is achieved 
through a series of convolutions with $3 \times 3$ kernels followed by average pooling (see Fig.~\ref{fig:encdec}). 

\noindent{\bf Bridge segment:} Unlike regular U-NETs~\cite{ronneberger2015u}, our bridge segment consists of three densely connected layers, instead of only one that is usually employed. This is designed to handle the multi-modal GloVe~\cite{pennington2014glove} embeddings of the token branch as well as the multi-faceted pictorial output.

\noindent{\bf Pictorial decoder branch:} The pictorial decoder branch of the network produces 44 image channels containing pixel-level output. 
We observed that adding a $1 \times 1$ convolution layer with a very high number of filters prior to the output layer increases the fidelity of these images. We use a $1 \times 1 \times 1500$ layer, a decision affected by available H/W resources (we train our network using an NVIDIA RTX A6000 GPGPU with $\approx49$GB VRAM). We attribute the benefits of this ``pixel-wise'' layer to increased capacity for a wider range of activated features directly before the final layer of the network. 

Both the pictorial/image output and the GloVe token embeddings use hyperbolic tangent ($tanh$) activations in the output layer. This choice forces output in the range of [-1..1], effectively controlling gradient flow. Remaining layers utilize Leaky ReLU~\cite{xu2015empirical} activations that also allow a small gradient when inactive to mitigate vanishing gradients.
Towards the same end, GloVe vectors from captioning output are also normalized to [-1..1], while the first layer after RGB input also rescales RGB values to [0..1]. As seen in Fig.~\ref{fig:encdec}, the encoder blocks of the network feature residual connections once again as a measure to diminish vanishing gradients. Residual connections are also featured, directly connecting encoder blocks to their respective same-dimensional decoder counterparts. 

\noindent{\bf Token/caption output branch:} The token/caption output branch departs from previous designs. In the past,  typical captioning networks used LSTMs~\cite{6795963} or GRUs~\cite{cho2014learning} featuring recurrent connections. ConvCAP~\cite{Aneja_2018_CVPR} maintained a purely convolutional approach. However, recurrence was still maintained because the network expected as input the previously regressed tokens in order to iteratively produce the next ones until forming a full sentence. This would be a prohibitive design choice for \ournet~ because it would immensely slow down the processing framerate. It is worth noting that although ConvCAP experimented with an early pairwise mechanism similar to transformer attention, this proved to have a very small positive impact upon the core ConvCAP architecture. Similarly, modern transformer architectures rely on positional encodings~\cite{vaswani2017attention} that vastly increase the complexity of the network by calculating attention scores for all possible output pairs.
Architectures like differential transformers~\cite{ye2024differentialtransformer} are now specifically built to tackle this problem by canceling out attention scores on ``noisy''/unimportant~\cite{ye2024differentialtransformer} token pairs, while research is being conducted on the attention heads themselves~\cite{zheng2024attention} and the roles they are called to accomplish.
In contrast to all this complexity, we follow a much more straight forward approach by eliminating the so called stop-word~\cite{stopwords} tokens, using GloVe as our token vector space. This helps with ordering the various responses of the network while training it. Lastly, residual connections from each token output to the next are incorporated in an effort to form encoding chains between tokens that build on previous tokens. 

It should be noted that the \ournet~ captioning branch features Layer- instead of Batch normalization between layers. 20\% dropout is applied after each token layer and gets conversely reduced as we go through from the first to the last token. Residual connections from one token to the next are scaled to 30\% of their initial magnitude to tackle the network producing the same token multiple times. Finally, output dimensions can vary depending on the task and computational resources. Overall, the full network we present outputs 44 heatmaps and 8 tokens: Heatmaps \#0 to \#16 encode 2D joints, \#17-\#28 PAFs~\cite{cao2017realtime}, heatmap \#29 depth, \#30-\#32 normals and \#33 to \#43 segmentation data.

\subsection{2D Human Pose Estimation (HPE)}
HPE can be formulated as a series of transformations of an input RGB image to heatmaps that are active only in the vicinity of a specific human body joint. Typical skeletons like the one used in COCO include 17 joints that contain the eyes, ears, shoulders, elbows, hands, hips, knees and ankles of a person.
\ournet~ can scale to more than COCO body joints. However, in an effort to constrain its already very broad scope, we opted to just focus on these 17 joints.

Being able to return 2D keypoints is very useful for detecting the presence of humans in the scene. However, connecting the dots of joints to form complete skeletons is equally important. The output of a 2D HPE module should not only be 2D joints but a list of humans with each joint associated to a specific skeleton. PAFs~\cite{cao2017realtime} were proposed to tackle this exact problem, indicating connected joint pairs. Therefore, we include regression of PAFs in outputs \#17-\#28 which are heatmaps that are active in the areas between neighboring joints and zero everywhere else. 

\subsection{Depth and Normals Estimation}
COCO17~\cite{lin2014microsoft}, much like virtually all non-synthetic, in-the-wild datasets, does not contain depth data for training. For this reason, we selected Depth Anything V2 (DAv2)~\cite{depth_anything_v2,yang2024depth} as a teacher model to generate them. DAv2 offers state of the art accuracy, and with it we can produce a depthmap for each RGB training sample of COCO17. Although DAv2 produces FP32 depth (similar to \ournet) its implementation's .png output is encoded as 8-bit values. We therefore modify it to encode depth as 16-bit. DAv2 features 2 families of models, extracting metric and normalized range depth results. It also features multiple model sizes. We use the biggest one publicly available (vitl 335M) and use normalized depth ranges since they produce rich details and are a superior choice in terms of visual fidelity~\cite{ranftl2020towards}.

Given a depth image \( D_I \) for each training sample, we compute the partial gradients 
\(\frac{\partial D_I}{\partial x}\) and \(\frac{\partial D_I}{\partial y}\) by applying Sobel filters 
along the \(x\) and \(y\) directions, respectively.
After performing gradient calculations, we compute the norm of the vectors, also adding a small value $\epsilon$ to avoid division by zero. We then compute the normal images $n_x, n_y$ and $n_z$. 

\begin{figure}[t]
  \centering
   \includegraphics[width=1.0\linewidth]{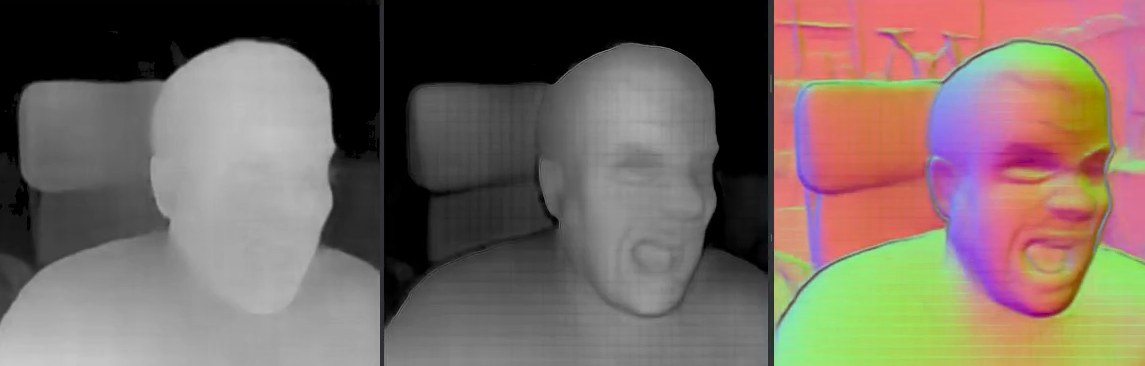} 
   \caption{ NN output normals (right) enforced on NN depth output (left) through our iterative algorithm can improve depth (mid).}
   \label{fig:improveddepth}
\end{figure}

At first glance, including normal data as an output of our NN seems counter-intuitive, since a network regressing depth already encodes normals in the depthmap. In practice, we observe that normal output has better fidelity compared to depth output. This is attributed to neighboring data that have a similar appearance in RGB maintaining similar $n_x, n_y$ and $n_z$ in their local neighborhoods, effectively reducing noise. We thus leverage this behavior, using the output normals to refine output depth as seen in Figure~\ref{fig:improveddepth}.

\noindent{\bf Iterative depth refinement}:
Assuming an output depth image $D_O$ and normal maps $N_x, N_y$ and $N_z$, we compute gradients in $x$ and $y$ directions by locally sampling depth:
$$G_x = \frac{\partial D_O}{\partial x}, \quad G_y = \frac{\partial D_O}{\partial y}. $$
We normalize gradients with their magnitude $M$ to obtain the normalized gradient maps 
$ G_x^{\text{nor}}$, $G_y^{\text{nor}}$, $G_z^{\text{nor}}$ 
and we calculate target normals and the gradients differences of the normalized depth map to find inconsistencies:
$$ \Delta_x = N_x - G_x^{\text{nor}}, \quad \Delta_y = N_y - G_y^{\text{nor}}, \quad \Delta_z = N_z - G_z^{\text{nor}}. $$
We finally update the improved output depth map \( D_O \) by minimizing the normal difference:
$$ D_O = D_O + \alpha \left( \Delta_x \cdot G_x + \Delta_y \cdot G_y + \Delta_z \right). $$

This process is executed iteratively for 35 iterations with $\alpha = 0.01$. We mask-out low depth (far) areas to skip calculations and respect our real-time target. As seen in Figure~\ref{fig:improveddepth}, this process produces noticeable improvements, sharpening resulting depthmaps and suppressing noisy values.

\subsection{Image Segmentation}
COCO 17 contains 183 segmentation label categories containing vehicles, animals, objects and other elements commonly visible in digital photographs. To perform per pixel segmentation, we need to arrange data in a way where, for each pixel, there is an associated output label. Encoding all possible discrete values into a single heatmap proved infeasible, since the neural network predictably lacks the capacity to produce such finely detailed activations at the segmentation output layer.
We thus need to decouple various labels as different heatmaps, similarly to how 2D joints are treated to make the task feasible for HPE.

However, using 183 separate $256 \times 256$ heatmaps, results in broadcasting $\approx 12$ million elements per training sample, leading to excessively long training times.
We attempt to encode all classes in one image yielding only 65K for transit to the GPGPU per training sample, and perform one hot encoding inside the loss function. 
Although this restores training times, with some expressivity loss, the capacity of the output layer doesn't allow this to work in practice. To overcome this problem, we group segmentation classes in 11 broad categories, namely Persons, Vehicles, Animals, Objects, Furniture, Appliances, Materials, Obstacles, Building, Nature and Text. Each label is assigned to its respective group yielding 11 segmentation images with the union of all visible classes. This balanced, divide \& conquer approach helps the network  successfully tackle a large number of labels while minimizing overheads.

\begin{figure}[t]
  \centering
   \includegraphics[width=1.0\linewidth]{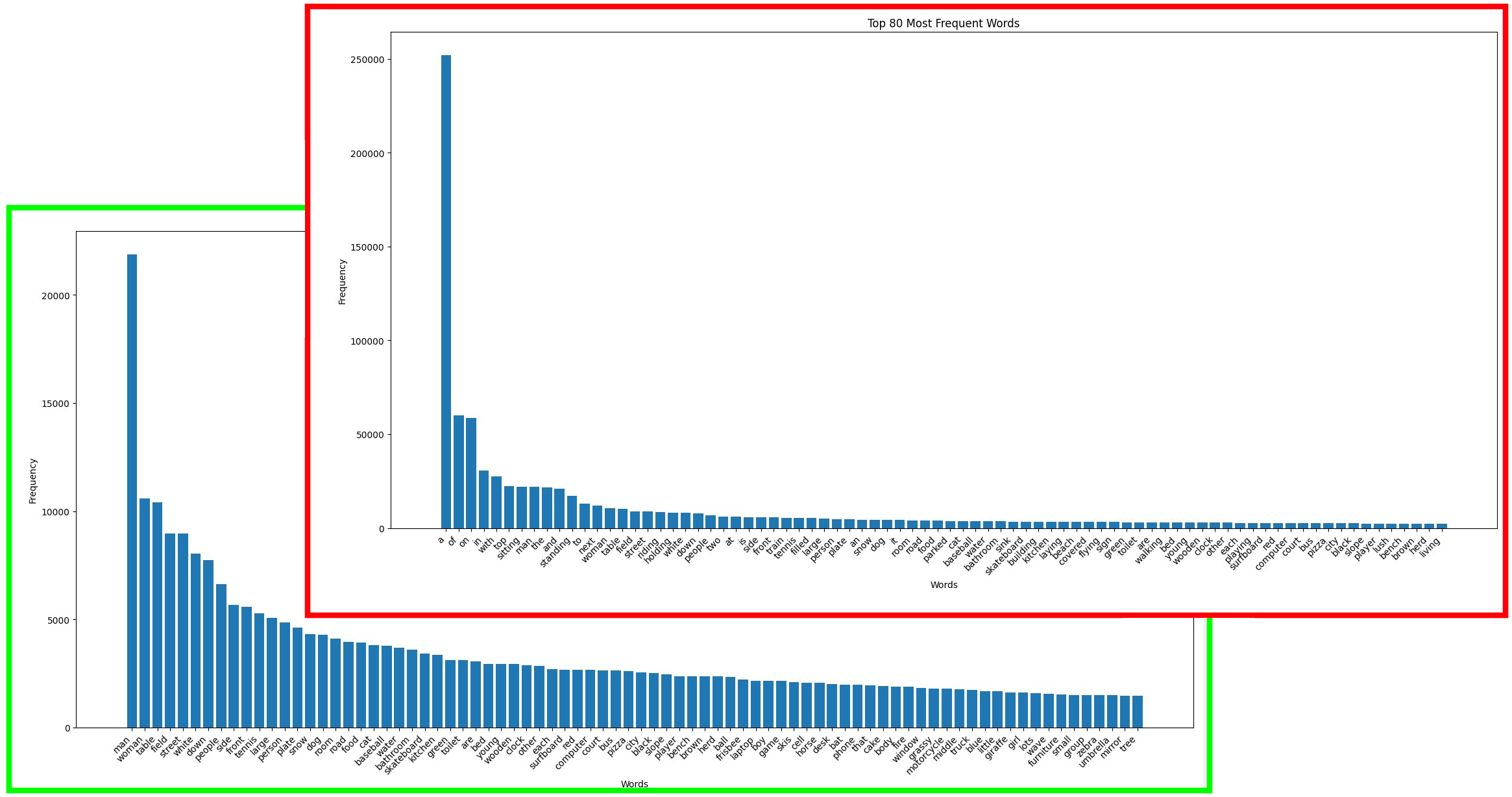} 
           \caption{  The frequency of tokens encountered while captioning an image follows a very heavy-tail distribution. {\bf Red:} Top-80 frequency words, when not using stop-words~\cite{stopwords}. Token `a' appears several orders of magnitude more than e.g. the token `cat'. This very heavy class imbalance negatively impacts training. {\bf Green:} After removing tokens: {\it `(',  `)', `.', `a', `an', `s', `of', `on', `and', `I', `in', `the', `is', `it', `at', `to', `with', `for' and `from'}, we get the second distribution (green frame) which is more balanced.
   }
   \label{fig:stopwords}
\end{figure}
\subsection{Multi-Label Captioning}
Captioning an image can be posed as selecting words that describe it and putting them in a sentence. However, as seen in Fig.~\ref{fig:stopwords} there's severe class imbalance among different tokens/labels. Works like ~\cite{multilabelcaptioning} highlight the intricate differences between multi-label, multi-instance, and regular captioning, motivating us to follow the former. To improve data distribution and conserve NN capacity, we remove stop-words (Fig.~\ref{fig:stopwords}). After limiting tokens to the 2048 most popular ones, initially, we naively attempted to perform multi-hot encoding for 8 output tokens, yielding $8 \times 2048$ captioning outputs for each sample. This approach however did not work since token order is very important and the task proves too difficult for a lightweight convolutional approach. 

Performing the union over these tokens and emitting a multi-hot 2048 class vector with 8 active labels per sample showed promise.  This training approach effectively removed token ordering, significantly simplifying the task. Despite performing class weighting based on the frequencies shown in Figure~\ref{fig:stopwords} however, the model tended to only produce results for a subset of tokens, and frequently made errors. This behavior is understandable: tokens like man, boy, woman, girl, person, and human have visually similar representations in images. Yet, despite their conceptual similarity, misclassifying a person as a man versus a cat incurs the same loss penalty, leading to problematic training.

To improve upon this, we employed Global Vectors (GloVe)~\cite{pennington2014glove} for word representation. GloVe offers different pre-trained databases. The vocabulary and source material for training can also vary ranging from Wikipedia, Gigaword, Common Crawl and Twitter/X for the pretrained models. We select GloVe 6B which is uncased, is trained on Wikipedia 2014 and Gigaword 5, with 6B tokens, a 400K vocabulary, and offers 50, 100, 200 or 300 dimensional embeddings for each word. Employing GloVe we lose the ability to perform a union (or sum) of words as a combined output. However, attempting to predict $8 \times 300$ captioning outputs, this time using a MSE loss, showed signs of promise even from very early experiments. Observed output artifacts included having multiple outputs regress similar embedding values. During runtime, to transform the GloVe 300D output back to words we follow three steps. First we threshold the norm of the 300D GloVe vector against zero, filtering it out if it is very close to zero. We then select the word token that has the highest cosine similarity with our output. As a final step we can perform a second threshold operation filtering out tokens with low cosine similarities.

In order to improve token sequences we introduced residual connections from each token to the next. We observed that this further increased token repetition. Using tanh output activations, normalizing GloVe embeddings to [-1..1], employing dropout, and layer instead of batch normalization, as well as scaling down the magnitude of the residual connections to 30\% mitigated this.
As reported by methods like ConvCAP~\cite{Aneja_2018_CVPR}, GloVe embeddings can be used as intermediate output supervision. Adding a final 2-layer densely connected tail (bottom of Figure~\ref{fig:model}) with sigmoid or softmax activations yields easy-to-use outputs for applications. This eliminates the need for extra computations, since thresholding the returned pseudo-probabilities is sufficient.

\subsection{Data Augmentation}
COCO17 contains 118K training samples. To enhance \ournet's generalization with this relatively limited amount of data and allow for a multi-epoch training regime without over-fitting, we employ aggressive data augmentation. The average sample dimension of COCO17 is close to VGA resolution. Our input and output images, however, have a $256 \times 256$ size.
We thus resize the images while respecting aspect ratios, and with black padding.
For each image we also apply a randomization scheme with the following properties. Each augmentation has an independent chance to be activated thus allowing multiple augmentations on an image. A sample has 45\% chance to be panned \& zoomed. It has 50\% chance to have its brightness/contrast altered, out of which, for half of the samples we perform uniform brightness/contrast changes across all channels, while the rest have R,G,B channels individually altered with up to a $\pm20\%$ change. There is a 15\% chance of Gaussian noise perturbation, 50\% chance of adding up to 10 burned pixels (up to 1/6553 corruption in $256 \times 256$ images). Finally, if a sample does not contain any persons there is a 1\% chance of completely replacing the image with random noise. This ensures not hurting human pose due to our limited training data while at the same time conditions the network to filter out erratic input. An important detail of the pan \& zoom augmentation is that it has a strict limit of up to 110\%. If zooming is left unchecked, captioned areas of the image might be cropped out. This can cause concept drift, with tokens like surfboards associated with e.g. the sea. Captions describe the whole image and erroneous zooming may e.g. crop-out visible surfing boards which occupy less space than the sea. Finally, if persons are present, our randomization ensures at least 30\% of existing joints of depicted persons visible inside the cropped image.

\subsection{Model Training}
We train our model using Keras 3.1.1 with Tensorflow 2.16.1 as a backend. We develop a data loader written from scratch in C and thoroughly profile and optimize it. We use a batch size of 30, corresponding to 30 threads concurrently executing on an AMD Ryzen Threadripper PRO 3955WX 16-Cores/32-Thread CPU. Back propagation is executed on an NVIDIA RTX A6000 GPU, where training occupies 47.2GB of the available 49.1GB VRAM. We also utilize a RAMFS to reduce I/O overhead.
Our model weights are in FP32 precision; however, to enhance training throughput, we use INT8 representations for images during their transfer to the GPU.
All training sessions start from weights following the Glorot/Xavier~\cite{glorot2010understanding} initialization. We use the ADAM~\cite{diederik2014adam} optimizer with a learning rate starting at $2.4 \times 10^{-4}$
and decaying linearly to $1.0 \times 10^{-4}$ after 100 epochs. In total, we allow 250 training epochs, with checkpoints~\cite{kerasCheckpoint} and early stopping~\cite{kerasEarlyStopping}  with 65 epoch patience.

Heatmap target outputs also have a decay schedule with joints having a Gaussian size of $23 \times 23$ during the start of training and decaying down to a target limit of $6 \times 6$. In a similar manner, PAFs start from a line width of 6 and decay down to 2. This reduction happens linearly with the size being reduced by one step every 20 epochs. This helps the NN both identify joints early on during training since their magnitude is comparable to other output modalities, but also ensures better localized joint responses for maturely trained models.
The loss function ${\cal L}$ we use is the following:
$$
{\cal L} = w \cdot \sum_{i} g_i \cdot \frac{1}{N} \sum_{j=1}^{N} \left( y_{\text{true, } i, j} - y_{\text{pred, } i, j} \right)^2.
$$
In the above equation, $i$ runs over 7 different terms. 
Each MSE term represents the mean squared error between the true and predicted values for the respective channels, weighted by its corresponding gain \(g_i\). We experimentally determined $g_i$ values that promote harder aspects of the problem by increasing their relative contribution to the total loss. These $7$ values are: 
    $g_G=10.0$ - for GloVe tokens (chan. 0-7), 
    $g_J=2.4$  - for joint heatmaps (chan. 0-16),
    $g_{\text{PAF}}=0.8$  - for PAFs (chan. 17-28),
    $g_D=1.0$ (for Depth - chan. 29),
    $g_N=1.0$  (for Normals - chan. 30-32), 
    $g_T=1.0$  (for Text  - chan. 33) and 
    $g_S=3.0$ (for Segmentation  - chan. $\geq$ 34). 
We use cosine similarity to monitor GloVe tokens. For image output we define {\bf Heatmap Distance Metric (HDM)}. Essentially, HDM measures the fraction of all predicted pixels within relative threshold ${\cal T}$ from their target value. In experiments (Tables~\ref{tab:ablation},~\ref{tab:quantitative}), ${\cal T}$ = 0.1 unless stated otherwise. Assuming that \( y_{\text{true}} \) and \( y_{\text{pred}} \) are true and predicted heatmaps in the [0..1] range, HDM is defined as:
$$ Y = \begin{cases} 
      1 & \text{if } |y_{\text{true}} - y_{\text{pred}}| \leq \mathcal{T},\\
      0 & \text{otherwise.} 
   \end{cases} $$
$$\text{HDM} = 
{\sum \mathbf{1}(Y = 1)}/{\text{(\#pixels)}}.$$
\section{Experiments} 
\label{sec:experiments}
\begin{table}[t]
\centering
\small
\resizebox{\columnwidth}{!}{
\begin{tabular}{|c | c r c c c c | c c c c r | c c|} 
 \hline
 \rotatebox{90}{\bf Exp. No.} & \rotatebox{90}{\bf Layers} & \rotatebox{90}{\bf M.Size} & \rotatebox{90}{\bf In.Dim.} & \rotatebox{90}{\bf Out.Dim.} & \rotatebox{90}{\bf Pix.wise} & \rotatebox{90}{\bf Res.con.} & \rotatebox{90}{\bf Pos.Dpth} & \rotatebox{90}{\bf Normals} & \rotatebox{90}{\bf PAFs}  & \rotatebox{90}{\bf Captions}  & \rotatebox{90}{\bf Class Grp.}& \rotatebox{90}{\bf HDM train} & \rotatebox{90}{\bf HDM val.} \\

 \hline\hline
 1 & 4 & 46M & 110 & 48  & - & - & \checkmark & -& - & - &-& 0.69 & 0.56\\ 
 \hline
 2 & 4 & 31M & 220 & 96  & - & - & \checkmark & -& - & - &-& 0.70 & 0.71\\ 
 \hline
 3 & 4 & 39M & 200 & 100  & - & - & \checkmark & -& - & - &-& 0.70 & 0.69\\ 
 \hline
 4 & 5 & 48M & 256 & 256  & - & - & \checkmark & -& - & - &-& 0.70 & 0.70\\ 
 \hline
 5 & 5 & 54M & 300 & 288  & - & - & \checkmark & -& - & - &-& 0.82 & {\bf 0.84}\\ 
 \hline 
 6 & 6 & 70M & 256 & 256  & - & - & \checkmark & -& - & - &-& 0.66 & 0.67\\ 
 \hline
 7 & 7 & 231M & 128 & 128  & - & - & \checkmark & -& - & - &-& 0.42 & 0.52\\ 
 \hline\hline
 8 & 7 & 136M & 420 & 384  & - & - & \checkmark & \checkmark & - & - & - & 0.75 & 0.76\\ 
 \hline
 9 & 7 & 278M & 420 & 384  & - & - & \checkmark & \checkmark & - & - & 19 & 0.74 & 0.77\\ 
 \hline
 10 & 8 & 401M & 420 & 256  & - & - & \checkmark & \checkmark & - & - & 19 & 0.74 & 0.77\\ 
 \hline
 11 & 7 & 315M & 256 & 256  & - & - & \checkmark & \checkmark& \checkmark &-& 3 & 0.56 & 0.61\\ 
 \hline
 12 & 7 & 316M & 256 & 256  & 1.6K& - & \checkmark & \checkmark& \checkmark &-& 3 & 0.62 & 0.67\\ 
 \hline
 13 & 7 & 172M & 256 & 256  & 1.6K& \checkmark & \checkmark& \checkmark& \checkmark &-& 3 & 0.66 & 0.72\\ 
 \hline
 14 & 7 & 213M & 256 & 256  & 1.7K& \checkmark & \checkmark & \checkmark& \checkmark &-& 2 & 0.53 & 0.59\\ 
 \hline
 \hline
 15 & 7 & 298M & 256 & 256  & 1.5K& {\bf \checkmark} & {\bf \checkmark} & {\bf \checkmark}& {\bf \checkmark} & {\bf \checkmark} &  11 & 0.64 & {\bf 0.72}\\ 
 \hline
 
 \hline 
\end{tabular}
} 
\caption{Ablation study for \ournet. We record enc/dec branch layer depth, model size (millions of weights), input and output dimensionality (images are rectangular so 256 is equivalent to 256x256), NN features such as pixelwise layer before output (Figure~\ref{fig:model}), residual connections in enc/dec blocks (Figure~\ref{fig:encdec}),  output modalities enabled in an experiment (PAFs, Normals, Class segmentation) and the training/validation HDM score for each configuration. {\bf Up:} Limiting output to joint+depth modalities we experimentally find that shallow networks perform better than more complex ones that over-fit. 
{\bf Down:} As more modalities are introduced the problem becomes harder and tensor sizes need to be reduced to maintain real-time performance.  The final row contains the \ournet~ 1-8-44 configuration. Its weights are increased compared to previous experiments due to the added captioning branch.  Each experiment requires $\approx$ 1 week to run. \label{tab:ablation} }
\end{table}
Our method addresses a broad range of computer vision sub-topics, making it the first to jointly tackle them. It uses DAv2~\cite{yang2024depth}, Detectron 2~\cite{wu2019detectron2}, DP-Text~\cite{ye2022dptext} and VisionGPT2~\cite{visiongpt2} as teacher models in the multi-teacher, single-student setting we adopt. It is trained on COCO17 with a training set of 118K samples, which is orders of magnitude smaller than the millions of samples training corpora of our teacher models. Furthermore, in order to provide diverse functionality in real-time, our model is less than half the size compared to the sum of weights of its teachers (Table~\ref{tab:advantage}). Last but not least, our method uses the same weights to simultaneously deal with all different output modalities, while each of its teachers only deals with a single problem. We thus expect our method to produce ``distilled'' output of reduced yet comparable accuracy, compared to its teachers.
As discussed in detail in the paragraph concluding Section~\ref{sec:related} the conceptually closest method, Sapiens, is not directly comparable to ours. In the absence of other methods combining the functionality we propose, we base our experimental comparisons primarily against our teacher models.

\subsection{Ablation Study}
We conduct a series of ablation experiments with different I/O sizes, different relative depths of the network,  gradually adding output modalities and broadening the size and scope of our NN until successfully accommodating pose, depth, normal, segmentation and captioning and reaching our target. These experiments are summarized in Table~\ref{tab:ablation}. 
We control network size by altering the number of filters in the enc/dec branches (Fig.~\ref{fig:model}). For example, as seen in rows \#6 and \#7 we can have a bigger NN that has enc/dec depth of 6 layers and a 3$\times$ larger NN handling half of the dimensionality of input and output in each dimension. As seen in experiment \#4, a 5-layer enc/dec architecture that tackles just pose and depth has excellent performance. Achieving a similar HDM metric with \#4 for all involved outputs, we need a substantially larger architecture with more capacity, like exp. \#13. Residual connections and the 1$\times$1 pixelwise layer, add representational capacity allowing stronger networks that tackle more modalities. Pixelwise layer width however, is limited by GPU VRAM. Adding a captioning branch, extra weights do not increase the image output branch. Caption features arriving at the bridge layer (Figure~\ref{fig:model}) however seem beneficial for the decoder branch. 

\begin{table}[t]
\centering
\small
\resizebox{\columnwidth}{!}{
\begin{tabular}{|l l | c c c c r|} 
 \hline
 \rotatebox{90}{{\bf Modality}} & \rotatebox{90}{{\bf Teacher}} & \rotatebox{90}{{\bf HDM 0.1}} & \rotatebox{90}{{\bf HDM 0.3}} & \rotatebox{90}{{\bf HDM 0.5}}& \rotatebox{90}{{\bf HDM 0.8}} & \rotatebox{90}{{\bf MSE}}  \\
 \hline\hline
 Joints      & COCO GT                                    & 0.95 & 0.95 & 0.95 & 0.95 & 2.22  \\ 
 \hline
 PAFs~\cite{cao2017realtime}        & COCO GT             & 0.96 & 0.96 & 0.96 & 0.96 & 1.45 \\ 
 \hline
 Depth       & DAv2~\cite{depth_anything_v2,yang2024depth}& 0.27 & 0.37 & 0.38 & 0.42 & 59.74 \\ 
 \hline
 Normals     & DAv2~\cite{depth_anything_v2,yang2024depth}& 0.53 & 0.64 & 0.68 & 0.75 & 58.79 \\  
 \hline
 Text Segm.  & DPText~\cite{ye2022dptext}                 & 0.94 & 0.94  & 0.94 & 0.94  & 1.29 \\
 \hline
 Class Segm. & Detectron 2~\cite{wu2019detectron2}        & 0.91 & 0.91  & 0.91 & 0.92  & 4.74 \\ 
 \hline
\end{tabular}
} 
\caption{ Quantitative results for specific NN output modalities using MSE and HDM for various $\mathcal{T}$ thresholds (0.1 - 0.8).\label{tab:quantitative} }
\end{table}
\subsection{Quantitative Experiments}
We evaluated our model on the COCO17 validation set, using ground truth data generated by our teacher models. Our quantitative metrics, summarized in Table~\ref{tab:quantitative}, reveal varying performance across different tasks. Depth estimation yielded the lowest accuracy, which aligns with its inherent complexity. In contrast to Joints, PAFs and class segmentations that have coarse ``on/off'' values and large inactive areas, depth maps return a response for all pixels. Furthermore even minor inaccuracies in depth predictions result in significantly larger error contributions, particularly when depth is assigned to an incorrect distance. Estimating object depth in in-the-wild images presents additional challenges due to an absence of reference scale, multiple viewpoints, or motion cues, as discussed in~\cite{xu2023unifying}.  Our depth teacher, DAv2~\cite{depth_anything_v2,yang2024depth} is trained on over 60M images, two orders of magnitude more than our training corpus, it has a larger size, and is dedicated to this single task as seen in Table~\ref{tab:advantage}. Normals perform better, with improved fidelity in areas with edges. We take advantage of this with our iterative algorithm (Figure~\ref{fig:improveddepth}). Class segmentation performs better and Joint/PAF estimation achieves the highest accuracy. We partly attribute this to the quality of human-annotated training data, since unsupervised model-generated labels may introduce inaccuracies. In captioning tasks, we use the cosine similarity metric achieving 0.59, 0.62, 0.64, 0.64, 0.53, 0.27, 0.08 and 0.01 scores for caption tokens 1 to 8. Overall we observe labels delivering a good generic image description.
Our .keras encoded model occupies 1.1 GB on disk. Although beyond the scope of this study, FP16 quantization could be beneficial due to our straightforward convolutional architecture. Benchmarking an unoptimized Python runtime on Intel i7-4790 / NVIDIA GTX 1070, it needs 7.5 GB of VRAM and reaches a 5.6 Hz refresh rate. On an AMD Ryzen 7 3800X / NVIDIA RTX 4080 SUPER, it utilizes 14 GB VRAM, achieving a real-time rate of 15.9 Hz.
\begin{table}[t]
\centering
\small
\resizebox{\columnwidth}{!}{
\begin{tabular}{|l | r | c c c c c c c|} 
 \hline
 \rotatebox{90}{{\bf Method}} &  \rotatebox{90}{{\bf M.Size}} & \rotatebox{90}{{\bf Joints}} & \rotatebox{90}{{\bf PAFs}} & \rotatebox{90}{{\bf Depth}} & \rotatebox{90}{{\bf Normals}} & \rotatebox{90}{{\bf ClassSeg.}} & \rotatebox{90}{{\bf Text Seg.}} & \rotatebox{90}{{\bf Captions}}  \\
 
 \hline\hline
  OpenPose~\cite{cao2017realtime}            & 28M   & \checkmark & \checkmark & - & - & -  & - & -  \\ 
 \hline
 DAv2~\cite{depth_anything_v2,yang2024depth} & 335M   & - & - & \checkmark  & \checkmark  & - & - & -  \\ 
 \hline
 Detectron2~\cite{wu2019detectron2}          & 63M   & - & - & - & - & \checkmark  & - & -  \\ 
 \hline
 DPText~\cite{ye2022dptext}                  & 44M   & - & - & - & - & - & \checkmark  & -  \\ 
 \hline
 VisionGPT2~\cite{visiongpt2}                & 239M   & - & - & - & - & - & - & \checkmark   \\ 
 \hline
 \hline
 {\bf \ournet~(Ours)} & 297M & \checkmark & \checkmark & \checkmark & \checkmark & \checkmark & \checkmark& \checkmark  \\ 
 \hline
\end{tabular}
} 
\caption{ We tackle a wide variety of tasks via a monolithic network using 41.88\% of the weights typically needed to perform the task. We use ~\cite{depth_anything_v2,wu2019detectron2,ye2022dptext,visiongpt2} as  teachers for \ournet \, training.  \label{tab:advantage} }
\end{table}
\begin{figure}[t]
  \centering
   \includegraphics[width=1.0\linewidth]{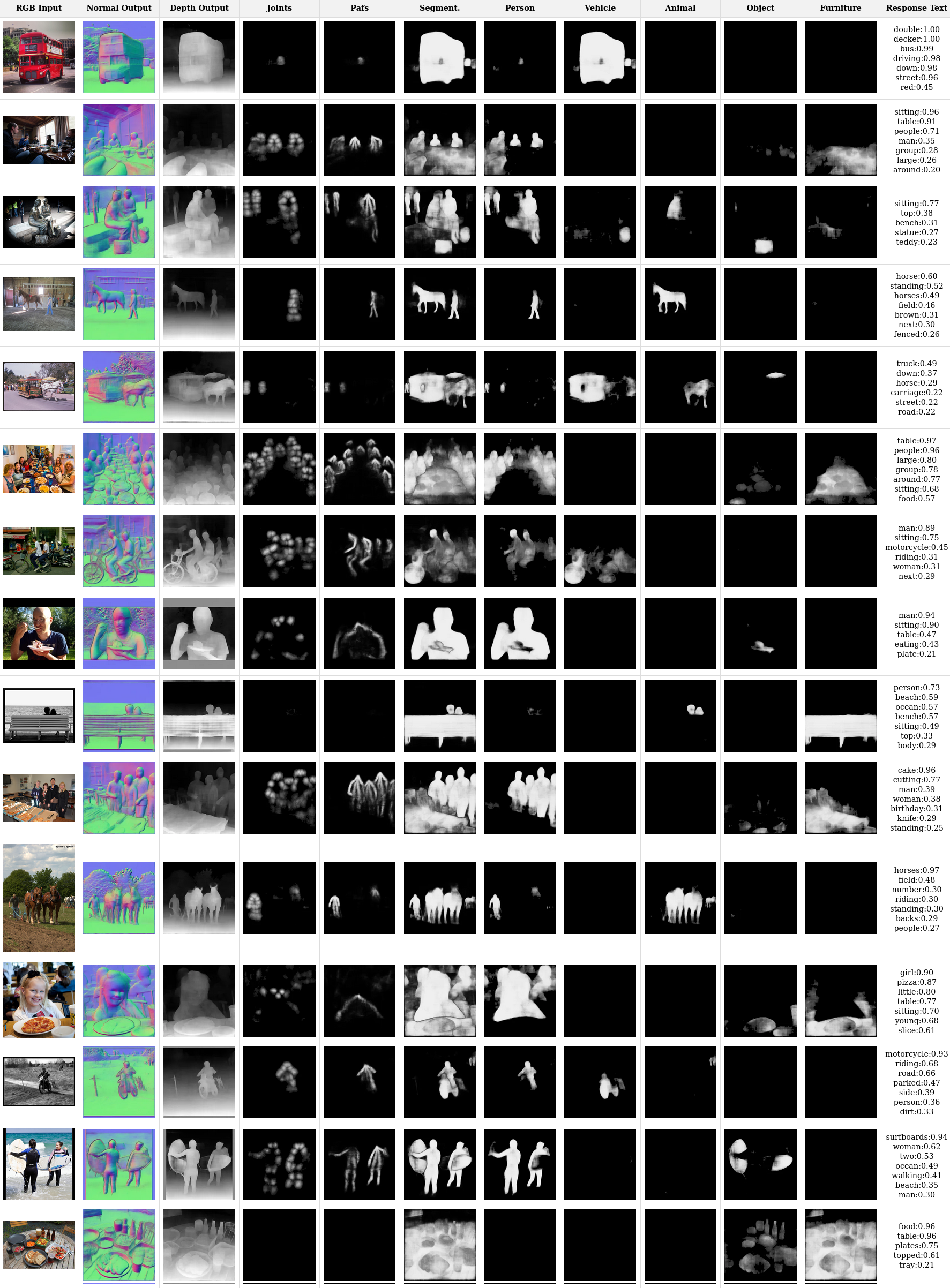} 
   \caption{ Qualitative results from COCO17 validation data showcasing various raw outputs of our model. Joints and PAFs contain the union of joint and PAF heatmaps, while column Segment. contains the union of columns 7-12 to showcase their synthesis.}
   \label{fig:qualitative}
\end{figure}
\subsection{Qualitative Experiments}
We test our method on a variety of input sources. We observe good generalization on the COCO17 validation set, showcased in Figure~\ref{fig:qualitative}. Since we aim at real-time applications in production environments, we experiment with Youtube videos featuring workers in production lines (see Fig.~\ref{fig:onecol}). Further samples of qualitative results are shown in Fig.~\ref{fig:qualitative}. We observe normal and depth estimation consistent with input, pose estimation results that follow the bodies even when wearing protective equipment, multi-token captioning output that describes the major visible components of the RGB image and class segmentation that, when observing a few objects / subjects, is very sharp. More qualitative results are provided in the supplementary material.
\section{Conclusions}
\label{sec:conclusion}
We presented the first work to jointly address depth, normal, pose, class segmentation and multi-label captioning in real-time. The proposed work offers an elegant and compact architecture that could have an important impact on the field due to its versatility, with the potential to become a staple along sparse architectures like YOLO and offering multi-modal vision capabilities in an efficient and scalable way. Future work includes using a mixture of experts supervision from multiple teacher models per modality and increased training corpus. To support research on the topic, our code will be made publicly available on github.

\section{Acknowledgments}
This work was co-funded by the European Union (EU - HE Magician – Grant Agreement 101120731) and the Hellenic Foundation for Research and Innovation (HFRI) under the ``1st Call for HFRI Research Projects to support Faculty members and Researchers and the procurement of high-cost research equipment'', project I.C.Humans, no 91. The authors also gratefully acknowledge the support of the framework of the National Recovery and Resilience Plan Greece 2.0, funded by the European Union – NextGenerationEU (project Greece4.0).

{
\small
\bibliographystyle{ieeenat_fullname}
\bibliography{main}
}

\end{document}